\documentclass{article}

    \usepackage[preprint]{neurips_2025}

\usepackage[utf8]{inputenc} 
\usepackage[T1]{fontenc}    
\usepackage{hyperref}       
\usepackage{url}            
\usepackage{booktabs}       
\usepackage{amsfonts}       
\usepackage{amsmath}
\usepackage{amssymb}
\usepackage{nicefrac}       
\usepackage{microtype}      
\usepackage{cleveref}

\usepackage{graphics}
\usepackage{epsfig}
\usepackage{wrapfig}

\usepackage{siunitx}
\usepackage{float}
\usepackage{bm}
\usepackage{enumitem}

\usepackage{caption}
\usepackage{subcaption}
\usepackage[table]{xcolor}
\usepackage{array}
\usepackage{multirow}
\usepackage{booktabs}
\usepackage{threeparttable}

\usepackage{pifont}

\usepackage{fontawesome5}

\definecolor{mygray}{gray}{.90}
\definecolor{mylgray}{gray}{.70}
\definecolor{reda}{RGB}{202,0,0}
\definecolor{redb}{RGB}{217,148,143}
\definecolor{lightred}{RGB}{125,0,0}
\definecolor{myyellow}{RGB}{190,144,0}
\definecolor{mygreen}{RGB}{0,136,51}
\definecolor{mylgreen}{RGB}{0,128,0}
\definecolor{myblue}{RGB}{0,102,204}

\usepackage{xspace}
\newcommand{\eg}{\emph{e.g.}\xspace} 
\newcommand{\ie}{\emph{i.e.}\xspace} 
\newcommand{\cmark}{\ding{51}}

\title{UniMRSeg: Unified Modality-Relax Segmentation via\\ Hierarchical Self-Supervised  Compensation}

\author{%
  \textbf{Xiaoqi Zhao}\textsuperscript{\textmd{1}}\quad
  \textbf{Youwei Pang}\textsuperscript{\textmd{2}}\quad
  \textbf{Chenyang Yu}\textsuperscript{\textmd{3}}\quad\\
  \textbf{Lihe Zhang}\textsuperscript{\textmd{3}}\quad
  \textbf{Huchuan Lu}\textsuperscript{\textmd{3}}\quad
  \textbf{Shijian Lu}\textsuperscript{\textmd{2}}\quad
  \textbf{Georges El Fakhri}\textsuperscript{\textmd{1}}\quad
  \textbf{Xiaofeng Liu}\textsuperscript{\textmd{1}}\quad\\
  \textsuperscript{\textmd{1}}Yale University, USA\\
  \textsuperscript{\textmd{2}} Nanyang Technological University, Singapore\\
  \textsuperscript{\textmd{3}}Dalian University of Technology, China\\
  \texttt{xiaoqi.zhao@yale.edu}\\
}

\begin{document}

\maketitle

\begin{abstract}
Multi-modal image segmentation faces real-world deployment challenges from incomplete/corrupted modalities degrading performance. While existing methods address training-inference modality gaps via specialized per-combination models, they introduce  high deployment costs by requiring exhaustive model subsets and model-modality matching. In this work, we propose a unified modality-relax segmentation network (UniMRSeg) through hierarchical self-supervised compensation (HSSC). Our approach hierarchically bridges representation gaps between complete and incomplete modalities across input, feature and output levels. 
First, we adopt modality reconstruction with the hybrid shuffled-masking augmentation, encouraging the model to learn the intrinsic modality characteristics and generate meaningful representations for missing modalities through cross-modal fusion.  
Next, modality-invariant contrastive learning implicitly compensates the feature space distance among incomplete-complete modality pairs.  Furthermore, the proposed lightweight reverse attention adapter explicitly compensates for the weak perceptual semantics in the frozen encoder. 
Last, UniMRSeg is fine-tuned under the hybrid consistency constraint to ensure stable prediction under all modality combinations without large performance fluctuations. 
Without bells and whistles, UniMRSeg significantly outperforms the state-of-the-art methods under diverse missing modality scenarios on MRI-based brain tumor segmentation, RGB-D semantic segmentation, RGB-D/T salient object segmentation. 
 The code will be released at \url{https://github.com/Xiaoqi-Zhao-DLUT/UniMRSeg}.
\end{abstract}
 \begin{figure}[!t]
	\centering
    	\setlength{\abovecaptionskip}{2pt}
	\includegraphics[width=\linewidth]{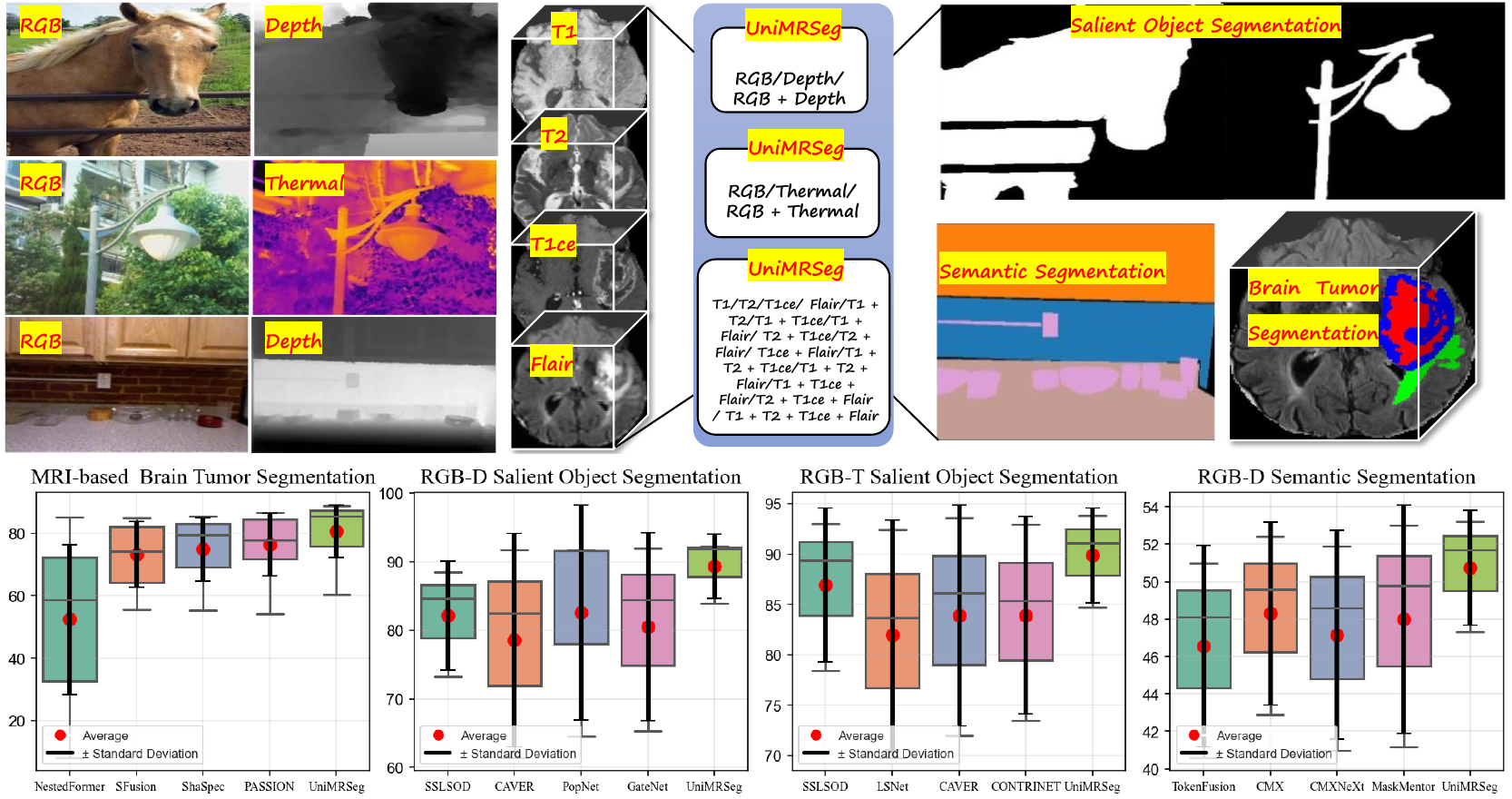}
	\caption{\textit{Top}: UniMRSeg has a unified framework and parameters within each segmentation task to handle 21 modality combinations (RGB-Depth: 3, RGB-Thermal: 3, MRI: 15). \textit{Bottom}: Box plots compare UniMRSeg with existing methods across four benchmarks, displaying average performance (red dots) and standard deviation (error bars).}
	\label{fig:unimrseg_task}
    \vspace{-7mm}
\end{figure}

\section{Introduction}
Visual multi-modal image segmentation has become a cornerstone in critical applications such as autonomous driving~\cite{Cityscapes}, medical diagnostics~\cite{brats2}, and robotics~\cite{robots1}, where complementary visual cues (\textit{e.g.}, RGB-D, MRI sequences) improve scene understanding.  
Advanced hybrid CNN-Transformer~\cite{UNETR,Eoformer}, global-local attention~\cite{global-local,DFormer}, dynamic convolution~\cite{HDFNet,MMFT}-based multi-modal fusion methods have achieved remarkable success under idealized settings with complete modalities.  
However, real-world scenarios often suffer from incomplete modality inputs due to sensor failures, low-quality data and clinical constraints. For example, although it is ideal to use four complementary MRI modalities—fluid-attenuated inversion recovery (Flair), contrast-enhanced T1-weighted (T1ce), T1-weighted (T1), and T2-weighted (T2)—for brain tumor diagnosis, variations in scanning protocols and patient conditions may limit the ability to obtain all MRI scans. 

Recent research on addressing missing modalities falls into two main challenges. \textit{\textbf{Firstly}},  
most methods~\cite{CMX,tokenfusion,CMNeXt,SSLSOD} focus on designing adaptable cross-modal interaction to amalgamate multi-modal features while preserving generalizable architectures for single-modal scenarios. 
However, during inference, diverse modality combinations require separate model parameters~\cite{CMX,tokenfusion} or independent encoder parameters~\cite{zeng2024missing,mmformer}, which not only increase resource consumption in practical deployment but also necessitate additional manual or automatic modality classification as a prerequisite. Although some works~\cite{M3AE,KD_brats} leverage the knowledge distillation from complete to incomplete modalities, they still demand multiple models for each modality subset, complicating clinical deployment. 
\textit{\textbf{Secondly}}, modality reconstruction-based methods~\cite{SSLSOD,zeng2024missing,M3AE,MaskMentor} aim to predict missing modality inputs to align features during training and inference. 
Since segmentation requires precise spatial features and boundary information, the pre-trained reconstruction model prioritizes global feature compression, resulting in insufficient feature representation. 
Therefore, it is difficult to directly inherit the ability to reduce the modality gap obtained by input-level~\cite{SSLSOD,M3AE} or feature-level~\cite{zeng2024missing,MaskMentor} reconstruction for downstream tasks (\textit{i.e.}, image segmentation). In particular, cascading low-quality reconstruction predictions~\cite{M3AE} as input to the segmentation network will increase error propagation and degrade performance.

\begin{wrapfigure}{r}{0.6\textwidth}  
  \centering
  \includegraphics[width=0.48\textwidth]{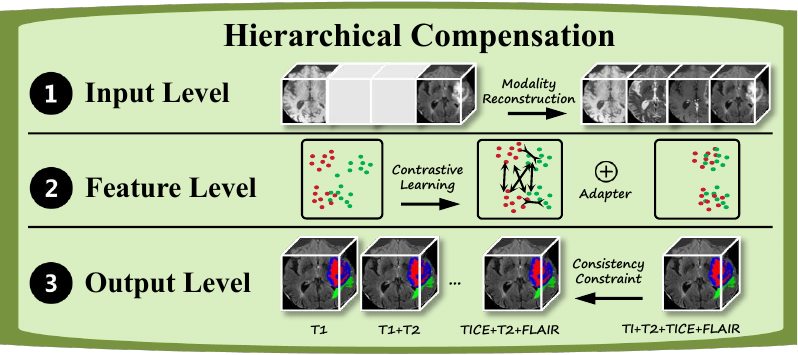}
  \caption{Illustration of the hierarchical compensation.}
 \label{fig:motivation}
\end{wrapfigure}

In this paper, we propose a unified modality-relax  segmentation framework (UniMRSeg). As shown in Fig.~\ref{fig:unimrseg_task}, UniMRSeg shares 100\% of its parameters across all possible modality input combinations in a given segmentation task. 
Since complete modality inputs typically yield the best prediction, our goal is to ensure that UniMRSeg, after training, can approach complete modality representation quality during inference with arbitrary modality inputs.  
To achieve this, we fully exploit the power of self-supervision~\cite{context1,MAE,SimCLR,wang2024progressively} in representation learning and propose a hierarchical compensation mechanism that operates at the input level, feature level, and output level, as shown in Fig.~\ref{fig:motivation}. 
\textit{\textbf{First}}, we adopt cross-modal reconstruction as a pretext task. Unlike previous methods~\cite{SSLSOD,Multimae} that either discard complete modalities or employ partial masking strategies, our approach simultaneously applies global and local masking mechanisms. This dual masking design enables the model to capture both fine-grained local patterns and holistic semantic representations from intra-modality and cross-modality interactions. Furthermore, we introduce a channel-wise modality shuffling operation that deliberately breaks the correspondence between input and reconstructed modalities. This operation implicitly formulates a modality classification task, compelling the model to disentangle modality-specific characteristics while eliminating dependence on modality category priors during inference. 
\textit{\textbf{Next}}, we leverage contrastive learning to finish the feature-level compensation. 
Specifically, we construct complete and randomly missing modalities from the same sample as positive pairs, and those from other samples as negative pairs. 
To enhance the inheritance of representation for  downstream segmentation tasks~\cite{mistretta2025cross,chng2024mask}, we jointly optimize the spatial distance metric and object segmentation at this stage to guide feature clustering in a direction that is beneficial for the results of the segmentation. 
Considering the inevitable prediction errors in the aforementioned pixel-level reconstruction and feature contrast learning, they may still limit the boundaries of the compensation mechanism. Inspired by adapter-based methods~\cite{adapter1,adapter2,adapter3}, we design a lightweight reverse attention adapter to explicitly compensate for weak perceptual semantics in the frozen encoder, while embedding a 3D Swin Transformer~\cite{VideoTransformer,3dTransformer} to capture high-response mutual attention patterns across modalities. 
By adding feature-level consistency constraints, we ensure that the adapter is aware of the partial representation defects inherent in any missing modal combination. 
\textit{\textbf{Last}}, UniMRSeg is fine-tuned by enforcing segmentation consistency constraints, and the knowledge presented by the complete modality at the output will be distilled to all missing modality combinations as supervision information. 
Through the hierarchical three-level compensation strategy, UniMRSeg achieves optimal average performance, the highest full-modality performance, and minimal performance variance, as shown in Fig.~\ref{fig:unimrseg_task}. 

Our main contributions can be summarized as follows: 
\begin{itemize}[leftmargin=*,itemsep=0em,topsep=0em,parsep=0em]
\item We propose a unified framework, \ie, UniMRSeg, with one set of parameters adapting to varying modality-missing scenarios for general multi-modal segmentation.
\item We integrate pixel-level modality reconstruction, feature-level contrastive learning, and prediction-level label distillation to construct a hierarchical annotation-free compensation mechanism, breaking through the previous isolated self-supervised research paradigm. 
\item Benefiting from the reverse attention adapter, UniMRSeg can explicitly obtain the compensation about difficult regions with weak perception of the complete modality, aligning missing and complete modality representations.  
\item Extensive comparisons conducted on brain tumor, salient object and semantic segmentation tasks in MRI (15 combinations), RGB-D (3 combinations), and RGB-T (3 combinations) modalities across 2D and 3D images within  medical and natural scenes, 
show that our method consistently achieves the best performance in all individual modality combinations while attaining superior average accuracy with minimal standard deviation. 
\end{itemize}

\section{Related Works}
\subsection{Incomplete Multi-modal Image Segmentation} 
There are three popular research patterns to handle missing modalities situations. 
\noindent\textit{\textbf{\uppercase\expandafter{\romannumeral1})
Generalizable Architectures.}} Some methods~\cite{tokenfusion,CMX,zhang2023delivering} aim to fully integrate multi-modal features while minimizing structural modifications in single-modal scenarios. CMX~\cite{CMX} introduces a cross-modal interaction attention mechanism that combines both the CNN and Transformer. Tokenfusion~\cite{tokenfusion} focuses on efficient token-based fusion, pruning multiple single-modal Transformer and repurposing the pruned units for multi-modal fusion. 
\noindent\textit{\textbf{\uppercase\expandafter{\romannumeral2})
Projection-based Methods.}} Hetero~\cite{Hetero} and HeMIS~\cite{HeMIS} perform arithmetic operations (\textit{e.g.}, averaging) in the projected space to obtain the final segmentation result. SFusion~\cite{SFusion} proposes a self-attention-based fusion block, where extracted features from available modalities are projected as tokens and processed through a self-attention layer to capture cross-modal relationships. 
\noindent\textit{\textbf{\uppercase\expandafter{\romannumeral3})
Reconstruction Strategies.}}  
M3AE~\cite{M3AE} refines the segmentation network by reusing the reconstructed modality images as inputs for fine-tuning. MaskMentor~\cite{MaskMentor} and SSLSOD~\cite{SSLSOD} treat modality reconstruction as a pre-training representation task, where missing and complete modalities share weights for joint training. Zeng \textit{et al.}~\cite{zeng2024missing} enforce consistency between missing modality feature reconstruction and complete modality features at the token level. 
Different from them, we aim to achieve a simple yet efficient architecture while completing the comprehensive compensation of multi-modal and multi-combinatorial representations with unified parameters by constructing multi-granularity self-supervised tasks that are not limited to pixel/token-level reconstruction.  

\subsection{Self-supervised Learning}
Self-supervised learning (SSL) has emerged as a powerful paradigm for learning representations without relying on manual annotations. 
Three principal SSL mechanisms dominate current research: 
\noindent\textit{\textbf{\uppercase\expandafter{\romannumeral1}) Mask-based SSL}} involves training models to predict missing or masked parts of an input. 
Early image inpainting works~\cite{pathak2016context,zhan2020self} evolved into modern masked autoencoders~\cite{MAE,wei2022masked} using high-ratio masking and asymmetric encoder-decoder architectures to various visual tasks, including image segmentation,  depth estimation, and low-level image restoration. 
\noindent\textit{\textbf{\uppercase\expandafter{\romannumeral2}) Contrastive learning}} encourages models to bring similar data points closer while distancing dissimilar ones.  
SimCLR~\cite{SimCLR} establishes data augmentation principles with momentum encoders, while MoCo series~\cite{MoCo,MoCov2} address negative sample scarcity through dynamic queues. CLIP~\cite{CLIP} leverages cross-modal contrastive learning to embed images and text into a unified semantic space, enabling zero-shot transfer capabilities and providing a universal representation foundation for multi-modal tasks like medical image retrieval and caption generation. 
\noindent\textit{\textbf{\uppercase\expandafter{\romannumeral3})  Knowledge distillation}}~\cite{KD1}  transfers supervision via teacher-student paradigms. In image segmentation, Chen \textit{et al.}~\cite{KD2} propose to normalize the activation map of each channel to obtain a soft probability map. For cross-modal scenarios, Gupta \textit{et al.}~\cite{KD3} transfer supervision from labeled RGB images to unlabeled depth and optical flow images. 
Existing studies usually develop these paradigms in isolation. We hope to take the incomplete multi-modal image segmentation task as an opportunity to effectively combine these three different levels of self-supervision techniques and demonstrate their synergy.

\section{Approach}
\label{sec:approach}

\noindent\textbf{Preliminaries.} 
In this section, we adopt the MRI-based brain tumor segmentation with the T1, T1ce, T2 and Flair modalities as the targeted task to describe the proposed multi-stage learning framework. 
Let $I \in \mathbb{R}^{1 \times H \times W \times T}$ be the input modality-specific sequence with $T$ slices for the model, where $1$, $H$, and $W$ are the channel, height, and width of the slice.
The final model generates the segmentation tensor $P \in \mathbb{R}^{N \times H \times W \times T}$, where each channel corresponds to a specific class and indicates the probability of each spatial-temporal location being assigned to that class.

\begin{figure*}[!t]
	\centering
    	\setlength{\abovecaptionskip}{2pt}
	\includegraphics[width=\linewidth]{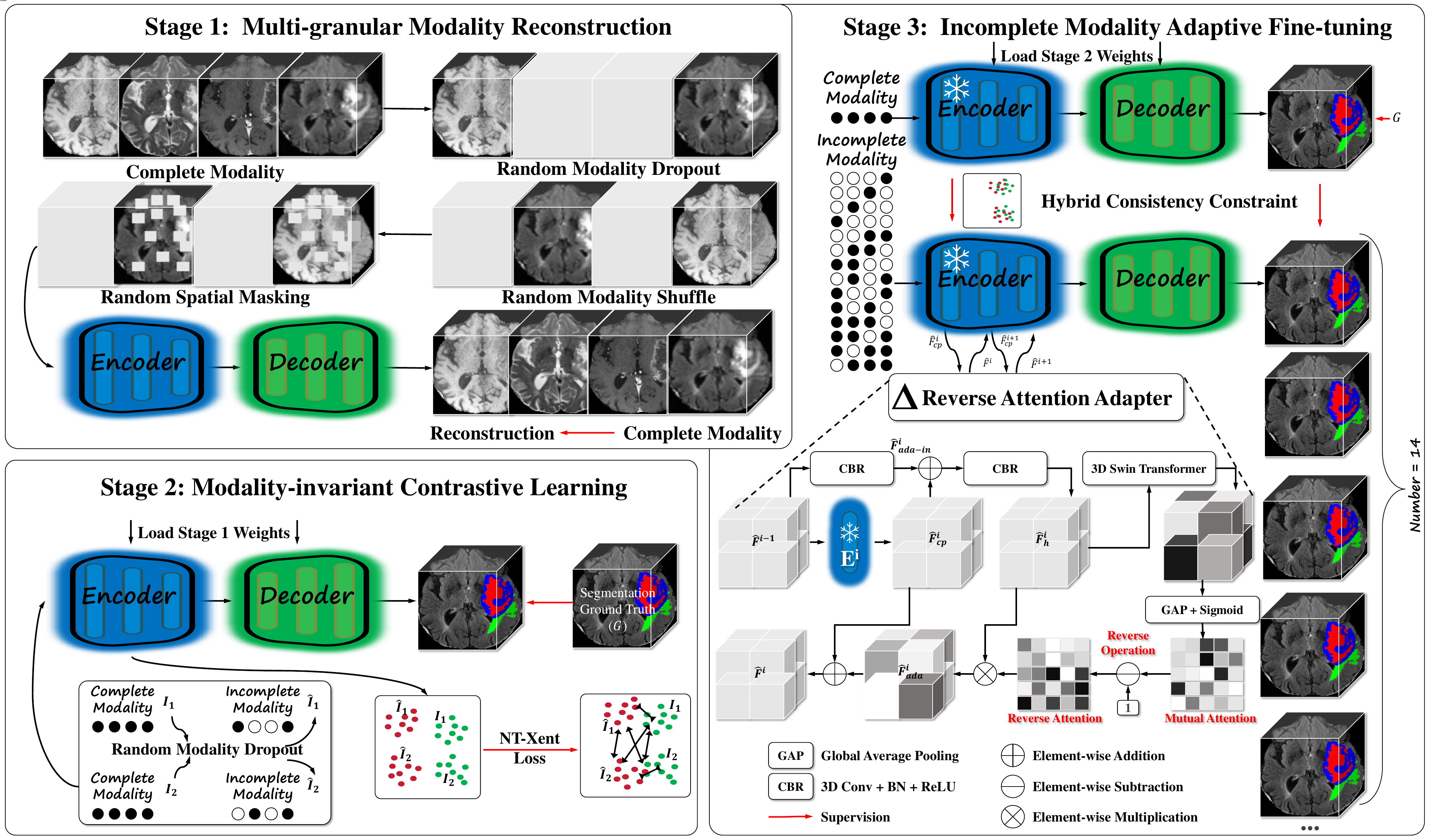}
	\caption{Multi-stage learning framework built on our encoder-decoder network.
    Stage 1: complete modality reconstruction based on multi-granular random perturbations.
    Stage 2: modality-invariant contrastive learning for enhancing incomplete-modality representation.
    Stage 3: incomplete modality adaptive fine-tuning via hybrid consistency constraints.}
	\label{fig:Architecture_comparison}
    \vspace{-5mm}
\end{figure*}

\noindent\textbf{Multi-stage Learning Framework.}
Our framework aims to bridge the performance gap between complete and incomplete multi-modal inputs through three-stage progressive learning, specifically designed to empower flexible handling of diverse missing modality scenarios while maintaining segmentation robustness. 
The overall pipeline is shown in Fig.~\ref{fig:Architecture_comparison}. The basic models in all stages follow a unified 3D U-Net-style~\cite{UNet} encoder-decoder structure and embed a 3D ASPP~\cite{ASPP} with dilation rates of [1, 6, 12, 18] into the high-level feature. 
The following sections elaborate on the technical specifics of each learning stage. 

\subsection{Multi-granular Modality Reconstruction}
\label{sec:stage1}

This stage emphasizes enhancing the representation capabilities under diverse potential input-side missing modality scenarios. 
Existing work~\cite{MAE} has demonstrated that effective data perturbations facilitate learning more robust representations. 
To encourage the model to mine implicit contextual relationships across modalities, this stage integrates three strategies, including \textit{modality dropout}, \textit{modality shuffle} and \textit{spatial masking}, to enable multi-granular information reconstruction.

\noindent\textbf{Data Perturbation.}
For the complete multi-modal input, we first apply the \textbf{random modality dropout} strategy to randomly discard some modalities with a 50\% probability and generate the output. 
And it also preserves at least one modality, thus retaining the fundamental information for overall reconstruction. 
Furthermore, we \textbf{randomly shuffle} the order of remaining modalities, which mitigates the model's reliance on fixed modality order and further decouples modality-agnostic representations from their inherent sequential dependencies in existing paradigms~\cite{MaskMentor,M3AE}. 
Additionally, the following \textbf{spatial masking} strategy randomly masks a portion of the input data, thus simulating missing effects within the sequences from available modalities.

\noindent\textbf{Data Reconstruction.}
We input these remaining perturbed samples into the 3D U-Net-based reconstruction network, and obtain the output through a ReLU function.
The normalized slices from the original complete modalities are used as reconstruction objectives based on the combination loss of L1 and SSIM~\cite{SSIM}, thus overall achieving self-supervised pre-training in the first stage. 

\subsection{Modality-invariant Contrastive Learning}
\label{sec:stage2}
The core idea behind this stage is to implicitly compensate for the feature biases introduced by missing modalities through contrastive learning.

\noindent\textbf{Representation Construction.} 
Without loss of generality, we consider a batch containing two input samples with complete modalities, \ie, $I_1$ and $I_2$.
By applying random modality dropout to each of them while ensuring at least one and at most three modalities remain active, we can obtain two extended samples ($\hat{I}_1$ and $\hat{I}_2$) with missing modalities. 
And then, we initialize the segmentation model by load the weights of the first stage except for the last output layer.
$\{I_1, \hat{I}_1, I_2, \hat{I}_2\}$ are then fed into the segmentation model to extract multi-level encoder features and generate the final predictions $\{P_1, \hat{P}_1, P_2, \hat{P}_2\}$, respectively.
After the global average pooling, those features are converted into four vector sets $\{f^i_1\}_{i=1}^5$, $\{\hat{f}_1\}_{i=1}^5$, $\{f_2\}_{i=1}^5$, and $\{\hat{f}_2\}_{i=1}^5$, respectively.
They are utilized for the following representation contrastive learning.

\noindent\textbf{Contrastive Learning.}
The aforementioned representation vectors are utilized to construct positive-negative sample relationships. 
And we introduce the NT-Xent loss~\cite{SimCLR} in each feature level to minimize distances between positive pairs and maximize separation of negative pairs. 
Specifically, for the $i^{th}$ level, positive pairs are from the same input sample and its augmented variant, \ie, $I^i_k$ and $\hat{I}_k$ ($k \in \{1, 2\}$).
Negative pairs are from distinct input sources, \eg, $(I_1, I_2)$ and $(I_1, \hat{I}_2)$.
Considering the vector set $\mathbf{f}^i = \{f^i_1, \hat{f}^i_1, f^i_2, \hat{f}^i_2\}$, we can calculate the NT-Xent loss as follows:
\begin{align}
    l^i(u,v)         & = -\log \frac{ \exp(\text{sim}(\mathbf{f}^i_u, \mathbf{f}^i_v) / \tau) }{ \sum_{k=1}^{2B} \mathbb{I}_{[k \neq i]} \exp(\text{sim}(\mathbf{f}^i_u, \mathbf{f}^i_k) / \tau) } \\
    L_\text{NT-Xent} & = \sum_{i=1}^5\sum_{k=1}^{B} \frac{ l^i(2k-1, 2k) + l^i(2k, 2k-1) }{2B \times 5}
    \label{equ:stage2_nt_xent_loss}
\end{align}
where $B=2$ and $\text{sim}$ are the batch size and the cosine similarity.
$\mathbb{I}_{[k \neq i]}$ is an indicator function that results in 1 when $k \neq i$ otherwise 0.
Such a contrastive learning encourages the model to learn modality-invariant representations, focusing on the underlying target semantic features rather than being biased by modality-specific characteristics.

\noindent\textbf{Segmentation Constraint.}
The common Dice loss $l_\text{Dice}$ is used to supervise the segmentation predictions as follows:
\begin{gather}
    L_\text{Dice} = \frac{1}{B} \sum_{k=1}^{B} \sum_{P \in \{P_k, \hat{P}_k\}} l_\text{Dice}(P, G_k)
    \label{equ:stage2_dice}
\end{gather}
It is worth noting that despite applying the contrastive learning, the segmentation constraint is also necessary to guide the encoder's representation learning.
This ensures the learned features are optimized and aligned for our ultimate objective, \ie, the segmentation task.

\subsection{Incomplete Modality Adaptive Fine-tuning}
\label{sec:stage3}

Due to potential errors in the reconstruction from the first stage and the positive-negative sample distance control in the second stage, we introduce the following adaptive fine-tuning guided by the complete modalities to dynamically compensate for these representation errors. 
As shown in Fig.~\ref{fig:Architecture_comparison}, our framework implements parallel pipelines to handle complete modality samples and their all potential incomplete counterparts (14 valid variants for 4 MRI modalities). 
The model directly generates the ideal intermediate features $\{F^i\}_{i=1}^{5}$ and segmentation prediction $P$ as the reference from complete modality samples via the forward propagation. 
And its prediction $P$ is supervised by the GT mask using the Dice loss. 
The incomplete samples additionally require targeted optimization through lightweight adapters $\{\mathbf{A}^i\}_{i=1}^{5}$, which progressively aligns their features $\{\hat{F}^i\}_{i=1}^{5}$ and prediction $\hat{P}$ with those references.
To maintain the representation capabilities pre-trained from the first two stages, we freeze the entire encoder during this stage and only fine-tune the decoder and adapters.

\noindent\textbf{Reverse Attention Adapter.} 
Taking the $i^{th}$ encoder stage as an example, we attach the adapter component to the frozen encoder stage $\mathbf{E}^i$. 
In the feature propagation path for incomplete multi-modal samples, the feature $\hat{F}^{i-1}$ from the previous encoder layer is first processed by $\mathbf{E}^i$ to obtain the base features $\hat{F}^{i}_{cp}$. 
Meanwhile, $\hat{F}^{i-1}$ also undergoes 3D convolutions and output the initial adaptive feature $\hat{F}^{i}_{ada-in}$, which is then integrated with $\hat{F}^{i}_{cp}$ via element-wise addition and sequential 3D convolutions.  
The generated feature $\hat{F}^{i}_{h}$ is fed into a 3D Swin Transformer~\cite{VideoSwin} block to establish global contextual correlations between the $\hat{F}^{i}_{cp}$ and $\hat{F}^{i}_{ada-in}$.
And then the global average pooling is applied across channel and sequence dimensions, followed by the sigmoid to generate the mutual attention. 
We hope to capture the difficult semantic parts that cannot be perceived by the first two stages and then compensate $\hat{F}^{i}_{cp}$.
Therefore, we apply the reverse operation to the mutual attention map and generate the reverse attention map.
And it is multiplied to $\hat{F}^{i}_{h}$ and generate the adapted feature $\hat{F}^i_{ada}$, which highlights differential information between modality-complete and -incomplete representations. 
Finally, the sum $\hat{F}^i$ of $\hat{F}^i_{ada}$ and $\hat{F}^i_{cp}$ replaces original $\hat{F}^i_{cp}$ as the real input to the subsequent process. The roles played by the above features need to be activated with the help of feature-level consistency constraints. 

\noindent\textbf{Hybrid Consistency Constraints.}
The consistency constraint process depicted in Fig.~\ref{fig:Architecture_comparison} involves the levels of encoder features and final predictions. 
For the \textbf{feature-level consistency}, we compare the intermediate features $\{F^i\}_{i=1}^{5}$ from modality-complete samples and $\{\hat{F}^i\}_{i=1}^{5}$ compensated by the adapter from modality-incomplete samples:
\begin{align}
    L_{fc} = \frac{1}{B} \sum_{k=1}^{B} \sum_m^{M} \frac{1}{5} \sum_{i=1}^5 \| F^i_k - \hat{F}^i_{k,m} \|_1
    \label{equ:stage3_feature_consistency}
\end{align}
where $M$ denotes the number of potential valid modality-incomplete samples and it is 14 in our MRI experiments.
Besides, the \textbf{prediction-level consistency} between the predicted segmentation maps from modality-complete and -incomplete samples can be formulated by:
\begin{align}
    L_{pc} = \frac{1}{B} \sum_{k=1}^{B} \sum_{m=1}^{M} l_\text{Dice}(P^i_k, \hat{P}^i_{k, m})
    \label{equ:stage3_prediction_consistency}
\end{align}
By accumulating these consistency constraints, the parameters of the decoder and adapters are jointly optimized to minimize the difference between the representations from modality-complete and -incomplete samples.

\begin{table*}[!t]
	\centering
    \caption{
Quantitative comparison of brain tumor segmentation on BraTS2020~\cite{BraTS}. $\uparrow$ and $\downarrow$ indicate that the larger scores and the smaller ones are better, respectively. \textbf{Note}: Official implementations are used when available. All methods are evaluated under a unified missing modality setting when possible. Some methods use different training data and lack released code, making exact re-training infeasible. Extended results under various settings are in the Appendix. 
	}
    \setlength{\abovecaptionskip}{2pt}
    \renewcommand{\arraystretch}{1.1}
    \renewcommand{\tabcolsep}{0.5mm}
	\resizebox{\linewidth}{!}{\begin{tabular}{cccc||ccccc|ccccc|ccccc}
	\hline
	\multicolumn{4}{c||}{} & \multicolumn{15}{c}{{Dice score(\%)}} \\ \cline{5-19} 
	\multicolumn{4}{c||}{\multirow{-2}{*}{{Modality}}} & \multicolumn{5}{c|}{{Whole Tumor (Whole)}} & \multicolumn{5}{c|}{{Tumor Core (Core)}} & \multicolumn{5}{c}{{Enhancing Tumor (Enhancing)}} \\
	\hline
	
	\multirow{2}{*}{Flair} & \multirow{2}{*}{{T1}} & \multirow{2}{*}{{T1ce}} & \multirow{2}{*}{{T2}} & {{NestedFormer}} & {{SFusion}} & {{ShaSpec}} & {{PASSION}} & {{UniMRSeg}} &  {{NestedFormer}} & {{SFusion}} & {{ShaSpec}} & {{PASSION}} & {{UniMRSeg}} &  {{NestedFormer}} & {{SFusion}} & {{ShaSpec}} & {{PASSION}} & {{UniMRSeg}}  \\
   &&&& {{~\cite{NestedFormer}}} & {{~\cite{SFusion}}} & {{~\cite{ShaSpec}}} & {{~\cite{PASSION}}} & {{}} &  {{~\cite{NestedFormer}}} & {{~\cite{SFusion}}} & {{~\cite{ShaSpec}}} & {{~\cite{PASSION}}} & {{}} &  {{~\cite{NestedFormer}}} & {{~\cite{SFusion}}} & {{~\cite{ShaSpec}}} & {{~\cite{PASSION}}} & {{}} \\
	\hline
	\textbf{$\circ$} & \textbf{$\circ$} & \textbf{$\circ$} & \textbf{$\bullet$} & 23.22&  58.54&  63.10&  64.83& \textbf{75.14} & 8.26& 37.20& 42.83& 50.36 & \textbf{56.31} & 2.31& 16.34& 22.15& 30.23 & \textbf{35.19} \\
	\textbf{$\circ$} & \textbf{$\circ$} & \textbf{$\bullet$} & \textbf{$\circ$} & 24.61&  60.82&  57.31&  60.92& \textbf{68.87} & 27.40& 45.19& 54.59& 54.78 & \textbf{79.20} & 35.85& 44.34& 64.93& 69.12 & \textbf{79.20} \\
	\textbf{$\circ$} & \textbf{$\bullet$} & \textbf{$\circ$} & \textbf{$\circ$} & 8.15&  55.69&  55.40&  54.10 & \textbf{60.41} & 6.77& 29.84& 34.26& 39.93 & \textbf{45.70} & 3.55& 14.13& 18.23& 22.13 & \textbf{28.70} \\
	\textbf{$\bullet$} & \textbf{$\circ$} & \textbf{$\circ$} & \textbf{$\circ$} & 61.18&  74.25&  79.54&  77.35 & \textbf{85.33} & 34.01& 43.23& 52.74& 50.27 & \textbf{61.81} & 29.84& 30.92& 37.40& 39.60 & \textbf{45.61} \\
	\textbf{$\circ$} & \textbf{$\circ$} & \textbf{$\bullet$} & \textbf{$\bullet$} & 47.22&  72.91&  70.60&  75.82 & \textbf{81.64} & 40.02& 71.82& 75.93& 80.35 & \textbf{82.01} & 49.13& 66.18& 74.23& 79.03 & \textbf{79.58} \\
	\textbf{$\circ$} & \textbf{$\bullet$} & \textbf{$\bullet$} & \textbf{$\circ$} & 40.87&  60.30&  67.58&  67.90 & \textbf{72.98} & 52.34& 75.62& 78.92& 79.85 & \textbf{84.60} & 56.87& 66.09& 77.13& 78.32 & \textbf{80.23} \\
	\textbf{$\bullet$} & \textbf{$\bullet$} & \textbf{$\circ$} & \textbf{$\circ$} & 70.54&  84.40&  80.23&  82.39 & \textbf{86.22} & 42.84& 64.31& 60.20& 60.55 & \textbf{67.90} & 33.61& 36.53& 48.08& 46.05 & \textbf{51.14} \\
	\textbf{$\circ$} & \textbf{$\bullet$} & \textbf{$\circ$} & \textbf{$\bullet$} & 22.77&  67.35&  72.40&  77.71 & \textbf{77.98} & 13.16& 42.92& 52.80& 51.60 & \textbf{59.71} & 7.26& 32.29& 36.02& 33.42 & \textbf{41.49} \\
	\textbf{$\bullet$} & \textbf{$\circ$} & \textbf{$\circ$} & \textbf{$\bullet$} & 59.18&  83.71&  85.31&  86.64 & \textbf{87.78} & 32.42& 55.62& 63.53& 52.87 & \textbf{70.00} & 29.40& 32.09& 43.52& 43.34 & \textbf{53.12} \\
	\textbf{$\bullet$} & \textbf{$\circ$} & \textbf{$\bullet$} & \textbf{$\circ$} & 58.57&  84.90&  82.80&  85.74 & \textbf{87.96} & 43.91& 73.30& 76.52& 78.94 & \textbf{85.71} & 54.05& 69.48& 77.28& 79.08 & \textbf{81.59} \\
	\textbf{$\bullet$} & \textbf{$\bullet$} & \textbf{$\bullet$} & \textbf{$\circ$} & 76.67&  80.52&  84.12&  85.42 & \textbf{86.91} & 63.12& 77.30& 79.34& 80.38 & \textbf{85.92} & 69.73& 74.36& 76.02& 78.12 & \textbf{81.28} \\
	\textbf{$\bullet$} & \textbf{$\bullet$} & \textbf{$\circ$} & \textbf{$\bullet$} & 73.00&  78.90&  81.13&  81.31 & \textbf{85.50} & 42.25& 63.62& 61.93& 63.95 & \textbf{67.43} & 32.44& 30.03& 39.89& 43.60 & \textbf{49.57} \\
	\textbf{$\bullet$} & \textbf{$\circ$} & \textbf{$\bullet$} & \textbf{$\bullet$} & 74.30&  80.65&  83.20&  83.52 & \textbf{87.51} & 60.04& 76.28& 78.93& 79.92 & \textbf{82.78} & 64.47& 70.02& 74.68& 79.09& \textbf{80.50} \\
	\textbf{$\circ$} & \textbf{$\bullet$} & \textbf{$\bullet$} & \textbf{$\bullet$} & 58.77&  70.52&  74.36&  76.31 & \textbf{76.57} & 60.32& 78.48& 80.65& 82.36 & \textbf{84.92} & 69.46& 67.18& 73.39& 76.04 & \textbf{77.15} \\
	\textbf{$\bullet$} & \textbf{$\bullet$} & \textbf{$\bullet$} & \textbf{$\bullet$} & 81.07&  84.92&  85.02&  85.90 & \textbf{88.74} & 67.02& 78.84& 84.26& 84.75 & \textbf{86.01} & 73.78& 72.09& 75.59& 80.72 & \textbf{82.18} \\
	\hline
	\multicolumn{4}{c||}{{Average $\uparrow$}} & 52.01 & 73.23 & 74.81 & 76.39 & \textbf{80.64} & 39.59 & 60.90 & 65.16 & 66.06 & \textbf{73.33} & 40.78 & 48.14 & 55.90 & 58.53 & \textbf{63.10} \\
	\hline
	\multicolumn{4}{c||}{{Std Dev $\downarrow$}} & 23.09 & 10.47 & 10.08 & 10.07 & \textbf{8.43} & 19.53 & 17.07 & 15.45 & 15.34 & \textbf{13.04} & 24.20 & 21.80 & 21.62 & 21.84 & \textbf{19.86} \\
	\hline
\end{tabular}}
	
	\label{tab:sota_compare}
\end{table*}

\begin{table*}[!t]
	\centering
        \caption{
    Quantitative comparison of segmentation performance across RGB, Depth and Thermal modalities. 
	}
    \setlength{\abovecaptionskip}{2pt}
    \renewcommand{\arraystretch}{1.1}
    \renewcommand{\tabcolsep}{0.5mm}
	\resizebox{\linewidth}{!}{
%

	\begin{tabular}{cc||ccccc|ccccc|ccccc}
	\hline
	\multicolumn{2}{c||}{{Modality}} & \multicolumn{5}{c|}{{RGB-D Salient Object Segmentation (STERE~\cite{STERE}) }} & \multicolumn{5}{c|}{{RGB-T Salient Object Segmentation (VT1000~\cite{VT1000}) }} & \multicolumn{5}{c}{{RGB-D Semantic Segmentation (SUN-RGBD~\cite{SUN-RGBD-37classes})}} \\
    \hline
		\multirow{2}{*}{RGB}  & \multirow{2}{*}{{Depth/Thermal}} & {{SSLSOD}} & {{CAVER}} & {{PopNet}} & {{GateNet}} & {{UniMRSeg}}  & {{SSLSOD}}&{{LSNet}} & {{CAVER}}  & {{CONTRINET}} & {{UniMRSeg}} & {{TokenFusion}} & {{CMX}} & {{CMXNeXt}} & {{MaskMentor}} & {{UniMRSeg}} \\
            
        & & {{~\cite{SSLSOD}}} & {{~\cite{CAVER}}} & {{~\cite{PopNet}}} & {{~\cite{GateNetv2}}} & {{}}  & {{~\cite{SSLSOD}}}&{{~\cite{LSNet}}} & {{~\cite{CAVER}}}  & {{~\cite{CONTRINET}}} & {{}} & {{~\cite{tokenfusion}}} & {{~\cite{CMX}}} & {{~\cite{CMNeXt}}} & {{~\cite{MaskMentor}}} & {{}}\\
		\hline
		\textbf{$\bullet$} & \textbf{$\circ$} & .846 & .825 & .916 & .844 & \textbf{.918} & .784 & .698 & .720 & .735 & \textbf{.847} & 48.1 & 49.6 & 48.6 & 49.8 & \textbf{51.7} \\
		\textbf{$\circ$} & \textbf{$\bullet$} & .732 & .614 & .645 & .653 & \textbf{.839} & .894 & .837 & .861 & .854 & \textbf{.911} & 40.6 & 42.9 & 41.0 & 41.2 & \textbf{47.3} \\
		\textbf{$\bullet$} & \textbf{$\bullet$} & .885 & .917 & .917 & .919 & \textbf{.923} & .930 & .924 & .936 & .929 & \textbf{.938} & 51.0 & 52.4 & 51.9 & 53.0 & \textbf{53.2} \\
		\hline
		\multicolumn{2}{c||}{{Average $\uparrow$}} & 
	.821 & .785 & .826 & .805 & \textbf{.893} & 
	.869 & .820 & .839 & .839 & \textbf{.899} & 
	46.6 & 48.3 & 47.2 & 48.0 & \textbf{50.7} \\
		\hline
		\multicolumn{2}{c||}{{Std Dev $\downarrow$}} & 
	.080 & .155 & .157 & .137 & \textbf{.047} & 
	.076 & .114 & .110 & .098 & \textbf{.047} & 
	5.4 & 4.9 & 5.6 & 6.1 & \textbf{3.1} \\
	\hline
\end{tabular}}
	\label{tab:sota_compare_sod_ss}
	  \vspace{-5mm}
\end{table*}

\section{Experiments}
\label{sec:Experiments}
\subsection{Datasets and Metrics}
In this work, we conduct experimental comparisons on four popular visual multi-modal image segmentation tasks to show the generalizability of the proposed method. \noindent\textit{\textbf{\uppercase\expandafter{\romannumeral1}) Brain Tumor Segmentation}}.
We follow most brain tumor segmentation methods~\cite{larrazabal2021orthogonal,NestedFormer,RFNet,SFusion} use the BraTS2020 dataset~\cite{BraTS}, which contains 369 images with four modality scans: T1ce, T1, T2, and Flair, along with three annotated regions: enhancing tumor, tumor core, and whole tumor, which are mutually inclusive.
The dataset is split into training (315), validation (17), and test (37) sets.  
\noindent\textit{\textbf{\uppercase\expandafter{\romannumeral2}) RGB-D Salient Object Segmentation}}.
We adopt the same training set as most methods~\cite{SSLSOD,GateNetv2,CAVER}, \ie, 1,485 samples from the NJUD~\cite{NJUD} and 700 samples from the NLPR~\cite{NLPR}.
The test dataset is STERE~\cite{STERE}, which contains 1,000 RGB and depth image pairs with complex scenes. 
\noindent\textit{\textbf{\uppercase\expandafter{\romannumeral3}) RGB-T Salient Object Segmentation}}.
We follow the setting of recent works~\cite{CAVER,LSNet,CONTRINET}, the training set only contains the 2,500 samples from VT5000~\cite{VT5000} and adopt the VT1000~\cite{VT1000} as the test set which contains 1,000 pairs of RGB-T images including more than 400 kinds of common objects collected in 10 types of scenes under different illumination conditions.  
\noindent\textit{\textbf{\uppercase\expandafter{\romannumeral4}) RGB-D Semantic Segmentation}}.
SUN-RGBD~\cite{SUN-RGBD-37classes} is the popular indoor scene benchmark with 37 classes. 
It contains 10,335 pairs of RGB-D images, with 5,285 pairs allocated for training and 5,050 for testing.
We introduce some widely used metrics in each field for fair evaluation, including Dice for brain tumor segmentation, S-measure~\cite{Smeasure} ($S_m$) for salient object segmentation and IoU for semantic segmentation. 

\subsection{Implementation Details}
\label{sec:Implementation}
All experiments are conducted on one NVIDIA A800 GPU.
We adopt basic image augmentation techniques to avoid overfitting, including random flipping, rotating and border clipping. 
We train the model for 300 epochs based on the AdamW optimizer~\cite{AdamW} with a warmup schedule, an initial learning rate of 0.0001, and a weight decay of 0.00001.
For RGB-D and RGB-T tasks, we concatenate the inputs into a 4-channel tensor to maintain the single-stream architecture. 
For fair comparison, we separately adopt ResNet-50~\cite{ResNet} and ConvNext-B~\cite{ConvNeXt} as the backbone for RGB-D/T salient object segmentation and RGB-D semantic segmentation, which are widely used in their respective fields.

\subsection{Evaluation}
\noindent\textbf{Quantitative Results}. 
We conduct thorough comparisons on all the four tasks, as shown in Tab.~\ref{tab:sota_compare} and Tab.~\ref{tab:sota_compare_sod_ss}. All these models are either directly tested for performance or retrained based on their publicly released code. 
To simulate missing modalities in practice and keep consistent with training, we fill missing modalities in MRI with zero pixel values and directly copy other existing modalities to complete the dual input of any missing modality in RGB-D and RGB-T tasks. In addition, we follow some methods with depth estimation functions ~\cite{SSLSOD,PopNet,MaskMentor} to cascade the intrinsic depth predictor as the input of the models for RGB-D salient object segmentation and RGB-D semantic segmentation tasks. 
It can be seen that our proposed UniMRSeg achieves excellent performance on all tasks with the best average performance and lowest standard deviation. These high accuracy, strong robustness and generalization capabilities across multiple modalities show its potential as a unified, reliable, and efficient segmentation framework.

\begin{figure*}[!t]
	\centering
    \setlength{\abovecaptionskip}{2pt}
	\begin{subfigure}[b]{\linewidth}  
		\centering  
		\includegraphics[width=0.7\linewidth]{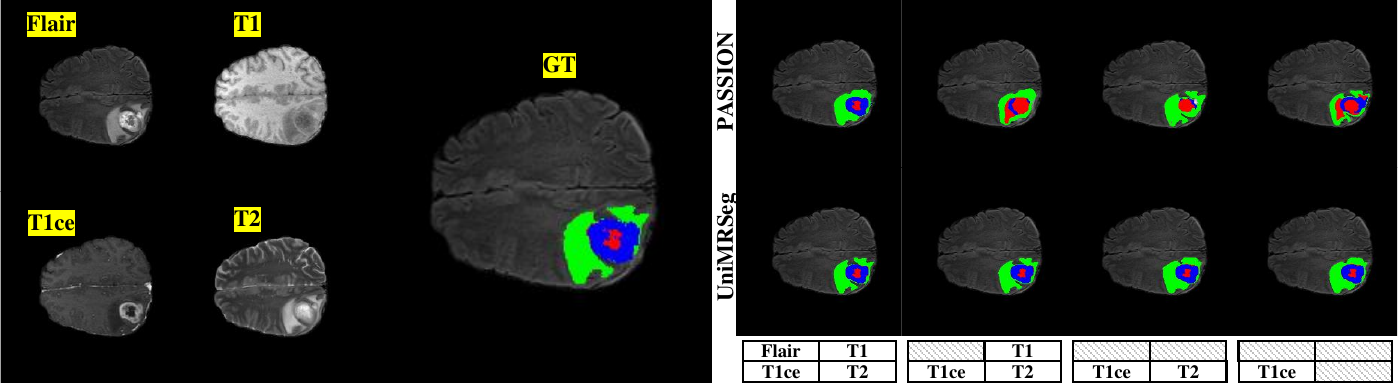}
		\caption{MRI-based Brain Tumor Segmentation}
        \label{fig:brain_tumor_segmentation_visualization}
	\end{subfigure}
	\par
	\begin{subfigure}[b]{0.3\linewidth}
		\centering
		\includegraphics[width=\linewidth]{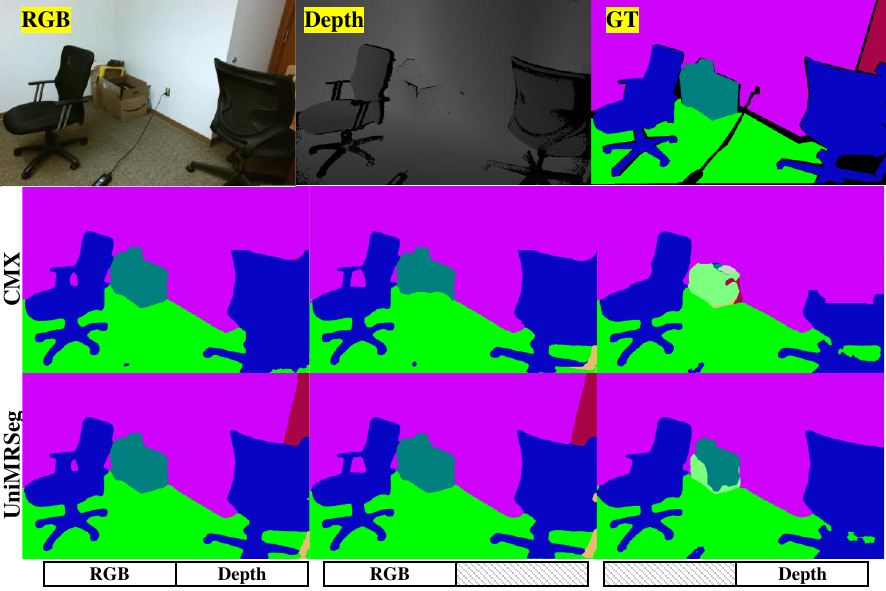}
		\caption{RGB-D Semantic Segmentation}
		\label{fig:rgbd_ss_visualizationv1}
	\end{subfigure}
	\hfill
	\begin{subfigure}[b]{0.3\linewidth}
		\centering
		\includegraphics[width=\linewidth]{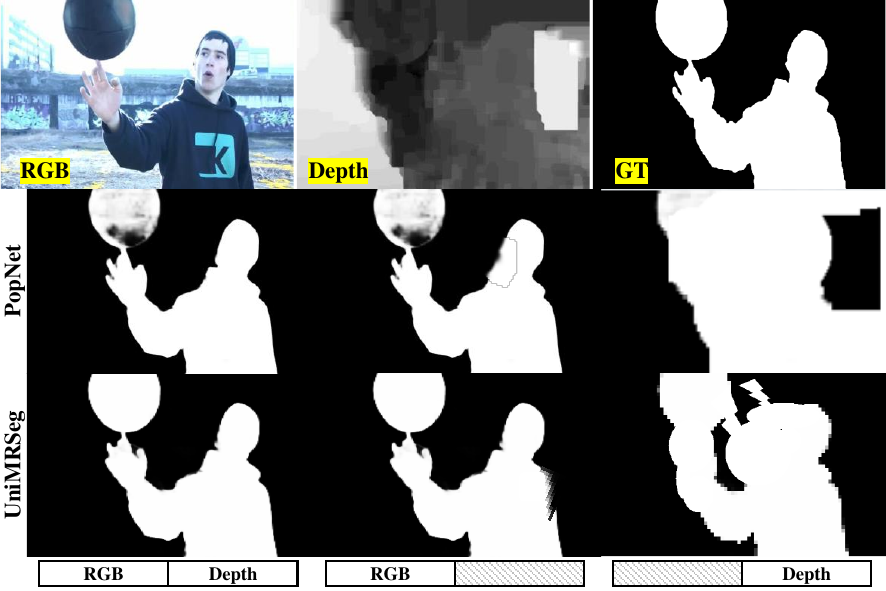}
		\caption{RGB-D Salient Object Segmentation}
		\label{fig:rgbd_sod_visualizationv1}
	\end{subfigure}
	\hfill
	\begin{subfigure}[b]{0.3\linewidth}
		\centering
		\includegraphics[width=\linewidth]{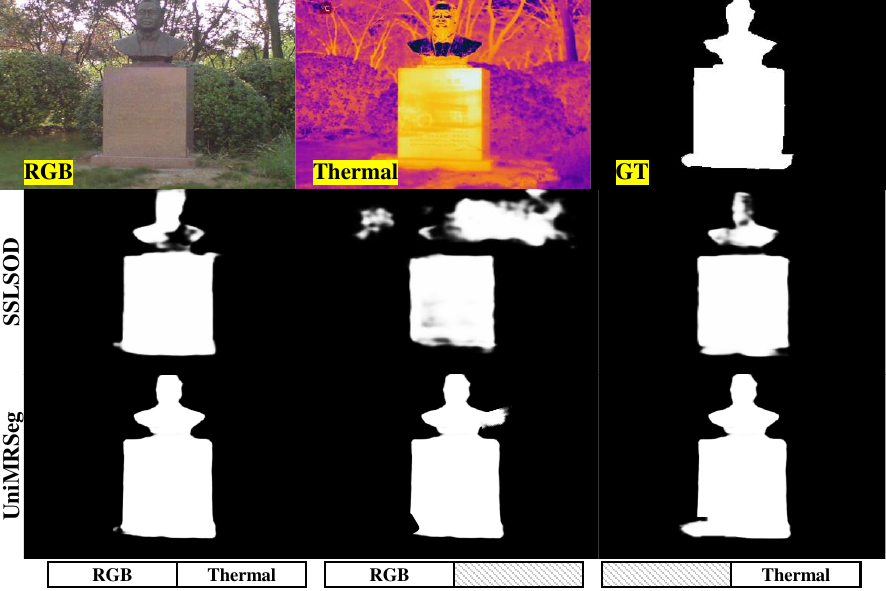}
		\caption{RGB-T Salient Object Segmentation}
		\label{fig:rgbt_sod_visualizationv1}
	\end{subfigure}
    
	\caption{
    Qualitative comparisons of predictions from different methods across different modality combinations.
    Best viewed on screen.
    }
	\label{fig:viusal_comparison}
    \vspace{-7mm}
\end{figure*}

\noindent\textbf{Qualitative Results}. 
In Fig.~\ref{fig:brain_tumor_segmentation_visualization}, PASSION~\cite{PASSION} exhibits significant prediction fluctuations for enhancing tumor (\textbf{\textcolor{blue}{blue}}) and tumor core (\textbf{\textcolor{red}{red}}) under missing modalities, while UniMRSeg maintains highly consistent predictions aligned with ground truth. T1 and Flair modalities notably degrade PASSION's boundary discrimination for whole tumor (\textbf{\textcolor{green}{green}}), revealing mutually exclusive fusion deficiencies. 
In Fig.~\ref{fig:rgbd_ss_visualizationv1}, UniMRSeg achieves precise door segmentation (\textbf{\textcolor{red}{red}}) by fusing color and geometric features across RGB/RGB-D modes, whereas CMX~\cite{CMX} fails completely with depth-only inputs. Leveraging spatial topology of chairs, UniMRSeg robustly segments blurred wall-adjacent boxes using pure depth data, demonstrating superior shape reasoning. 
Fig.~\ref{fig:rgbd_sod_visualizationv1} shows PopNet~\cite{PopNet} produces identical spherical predictions in RGB/RGB-D modes via embedded depth prediction, but degenerates to depth map binarization without RGB. In contrast, UniMRSeg reconstructs RGB semantics from depth and fuses cross-modal features to accurately segment humans and spheres. 
Fig.~\ref{fig:rgbt_sod_visualizationv1} verifies thermal modality's advantage in resolving RGB's color-similarity-induced sticking issues. Remarkably, UniMRSeg achieves complete target segmentation using thermal data alone.

\subsection{Ablation Study}\label{sec:Ablation}
In this section, we show the effectiveness of each component on the brain tumor segmentation task with the most complex modality combination. The baseline model is 3D-UNet without any pre-training.
 
\noindent\textbf{Each Components}. 
As shown in Tab.~\ref{tab:ablation_study}, the three data perturbations in stage 1 demonstrate the effectiveness of learning intra- and inter-modal features for multi-granular reconstruction. Their accumulation during pre-training leads to over 14.7\% performance gain over the baseline in subsequent segmentation.
Stage 2 highlights the importance of jointly training the segmentation decoder and the encoder guided by spatial distance clustering, which promote each other.
Stage 3 further optimizes cross-modal alignment through reverse attention adapters and prediction-level consistency constraints. The latter alone surpasses the stage 2 decoder by 12.0\%, showing the importance of enforcing prediction consistency across modalities. Gains in each stage are consistently and gradually improved, with the final model outperforming the baseline by over 44.6\% on average.

\begin{figure}[!t]
\centering
\begin{minipage}{0.50\linewidth}
    \centering
        \captionof{table}{Ablation study of each component.}
    \resizebox{\linewidth}{!}{	\begin{tabular}{l||ccc}
	\hline                                                                                   & \multicolumn{3}{c}{{Average Dice score(\%)}}                                                                                                                                                                                                                                                                                                                                                                                                                                                                                                                                                                                         \\ \cline{2-4} 
	\multicolumn{1}{l||}{\multirow{-2}{*}{{Models}}}                                                                                               & {{Whole}}                                                                                                                                                                        & {{Core}}                                                                                                                                                                         & {{Enhancing}}                                                                                                                                                                     \\
	\hline
	{{Baseline}}  & {{63.31}} & {{51.60}} & {{38.40}} \\
	\hline
	\multicolumn{4}{l}{\textit{Stage 1: Multi-granular Modality Reconstruction (Sec.~\ref{sec:stage1})}} \\   
		\hline 
	{{+ Random Modality Dropout}}  & {{66.98}} & {{55.47}} & {{42.25}}  \\
	{{+ Random Modality Shuffle}}  & {{67.78}} & {{56.85}} & {{44.17}}  \\
		\rowcolor[RGB]{235,235,250} 
	{{+ Random Spatial Masking}}  & {{69.35}} & {{59.89}} & {{47.12}}  \\
		\hline
	\multicolumn{4}{l}{\textit{Stage 2: Modality-invariant Contrastive Learning (Sec.~\ref{sec:stage2})}} \\ 
				\hline
	{{+ Contrastive Learning (Encoder)}}  & {{72.45}} & {{64.02}} & {{51.45}}  \\
			\rowcolor[RGB]{235,235,250} 
	{{+ Segmentation Constraint (Decoder)}}  & {{74.53}} & {{65.25}} & {{53.97}}  \\
	\hline
	\multicolumn{4}{l}{\textit{Stage 3: Incomplete Modality Adaptive Fine-tuning (Sec.~\ref{sec:stage3})}} \\ 
    \hline
    {{+ Feature-Level Consistency (Adapter)}}  & {{78.12}} & {{69.38}} & {{59.25}}  \\
    	\rowcolor[RGB]{235,235,250} 
    {{+ Prediction-Level Consistency (Segmentation)}}  & {{80.64}} & {{73.33}} & {{63.10}}  \\
	\hline
    \multicolumn{4}{l}{\textit{Three-Stage Design vs. Unified Single Stage}} \\ 
    \hline
     {{Three-Stage Design}}  & {{80.64}} & {{73.33}} & {{63.10}}  \\
          {{Unified Single Stage}}  & 20.32	&13.67	&10.03  \\
     \hline
\end{tabular}

}
    \label{tab:ablation_study}
  \end{minipage}
\hfill
\begin{minipage}{0.44\linewidth}
    \centering
   \begin{minipage}{\linewidth}
    \centering
        \captionof{table}{Ablation study of the reverse attention adapter.}
    \resizebox{\linewidth}{!}{	\begin{tabular}{l||ccc}
	\hline                                                                                   & \multicolumn{3}{c}{{Average Dice score(\%)}}                                                                                                                                                                                                                                                                                                                                                                                                                                                                                                                                                                                         \\ \cline{2-4} 
	\multicolumn{1}{l||}{\multirow{-2}{*}{{Models}}}                                                                                               & {{Whole}}                                                                                                                                                                        & {{Core}}                                                                                                                                                                         & {{Enhancing}}                                                                                                                                                                     \\
	\hline
		{{UniMRSeg}}  & {{80.64}} & {{73.33}} & {{63.10}} \\
	\hline
	{{w/o Rervese Attention}}  & {{78.28}} & {{69.26}} & {{60.45}} \\
    {{w/o Rervese Attention+Mutual Attention}}  & {{78.12}} & {{68.98}} & {{60.23}} \\
    {{w/o 3D Swin Transformer+Rervese Attention+Mutual Attention}}  & {{77.44}} & {{68.74}} & {{59.24}}  \\
    	\hline
{{Fine-tune Encoder in Stage 3}}  & {{77.05}} & {{68.15}} & {{58.20}} 
    \\

	\hline
\end{tabular}

}
    \label{tab:ablation_study_RAA}
  \end{minipage}
    \vspace{0.5cm} 
 \begin{minipage}{\linewidth}
    \centering
        \captionof{table}{Evaluation of different compensation levels.}
    \resizebox{\linewidth}{!}{

\begin{tabular}{ ccc||ccc }
	\hline
	Input Level                  & Feature Level                    & Output Level                              & \multicolumn{3}{c}{{Average Dice score(\%)}}                             \\ 
	(Stage 1)                    & (Stage 2 and Adapter) & (Segmentation Consistency) & {{Whole}}                                    & {{Core}}  & {{Enhancing}} \\
	\hline
	\cmark                       & -                                & -                                         & {{69.35}}                                    & {{59.89}} & {{47.12}}     \\ 
	-                            & \cmark                           & -                                         & {{72.46}}                                    & {{62.30}} & {{50.92}}     \\ 
	-                            & -                                & \cmark                                    & {{69.47}}                                    & {{57.25}} & {{48.31}}     \\          
	\hline
	\cmark                       & \cmark                           & -                                         & {{78.12}}                                    & {{69.38}} & {{59.25}}     \\
	\cmark                       & -                                & \cmark                                    & {{74.45}}                                    & {{66.40}} & {{54.38}}     \\
	-                            & \cmark                           & \cmark                                    & {{75.65}}                                    & {{67.48}} & {{54.95}}     \\
	\hline
	\rowcolor[RGB]{235,235,250}
	\cmark                       & \cmark                           & \cmark                                    & {{80.64}}                                    & {{73.33}} & {{63.10}}     \\
	\hline
	\multicolumn{1}{l}{Baseline} &                                  &                                           & {{63.31}}                                    & {{51.60}} & {{38.40}}     \\
    \hline
\end{tabular}
}
    \label{tab:ablation_study_three_level}
  \end{minipage}
\end{minipage}
\vspace{-5mm}
\end{figure}

\begin{figure}[!t]
	\centering
    	\setlength{\abovecaptionskip}{2pt}
	\begin{subfigure}[b]{0.23\textwidth}
		\includegraphics[width=\textwidth]{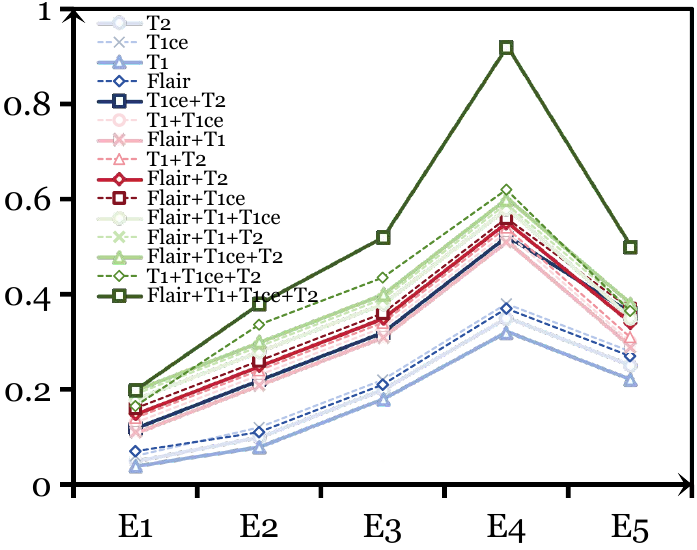} 
		\caption{Baseline}
		\label{fig:gap_value_1}
	\end{subfigure}
	\hfill 
	\begin{subfigure}[b]{0.23\textwidth}
		\includegraphics[width=\textwidth]{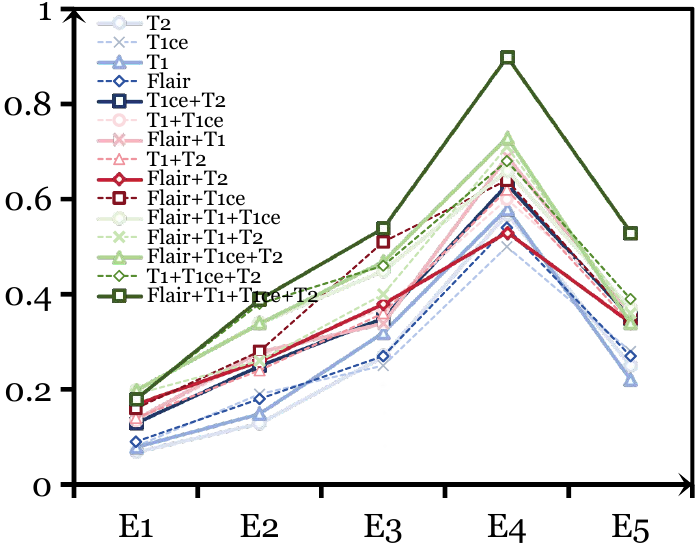} 
		\caption{Stage 1}
		\label{fig:gap_value_2}
	\end{subfigure}
	\hfill 
	\begin{subfigure}[b]{0.23\textwidth}
		\includegraphics[width=\textwidth]{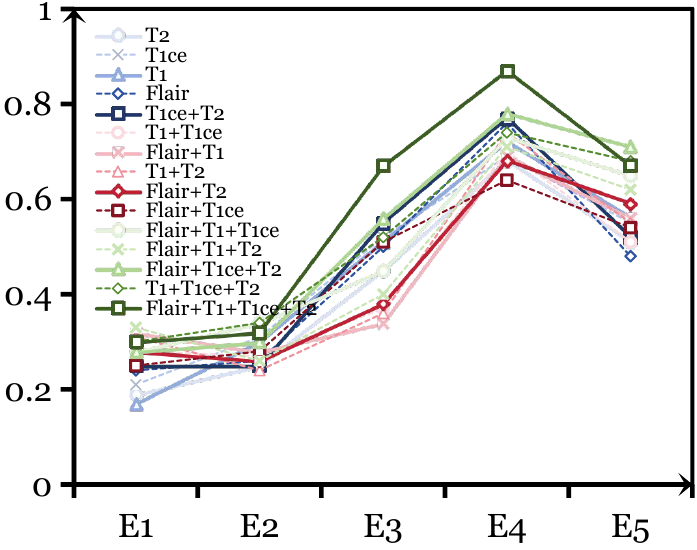} 
		\caption{Stage 2}
		\label{fig:gap_value_3}
	\end{subfigure}
	\hfill 
	\begin{subfigure}[b]{0.23\textwidth}
		\includegraphics[width=\textwidth]{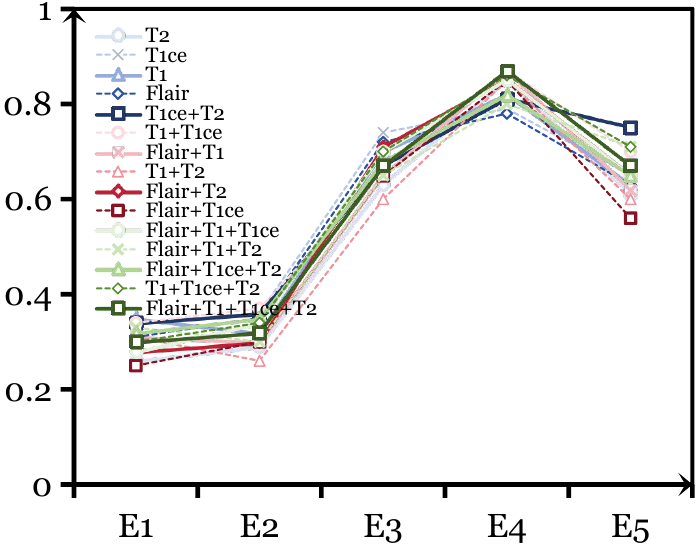} 
		\caption{Stage 3}
		\label{fig:gap_value_4}
	\end{subfigure}
	\caption{Average response value of feature maps of each encoding layer under different modal combinations at different stages.}
	\label{fig:gap_value}
    \vspace{-5mm}
\end{figure}

\noindent\textbf{Three-Stage Design vs. Unified Single Stage}. 
UniMRSeg is built upon a self-supervised pretraining framework, with training initialized from scratch starting at Stage 1. For each task (e.g., BraTS 2020), self-supervised training is conducted solely on the task’s own training set, rather than relying on externally labeled datasets such as ImageNet-pretrained weights for initialization.  
Our three-stage design includes: 1) Pretrain both the encoder and decoder through arbitrary modality reconstruction. 2) Pretrain the encoder with contrastive learning. The segmentation task here is only used to guide the contrastive objective, not as a final goal. 3) Perform the downstream segmentation task. 
In this way, our three-stage design explicitly aims to reduce the representation gap between complete and incomplete modalities in the encoder-decoder space during the final segmentation stage. If all these tasks are trained jointly in a single-stage model, it would fall into the scope of multi-task learning, rather than self-supervised pretraining.

In multi-task learning, the goal is to leverage multiple complementary clues for collaborative learning. These clues can come from the data level~\cite{SAM,Seggpt,CDCU-Spider}, or from structural supervision types~\cite{MTL1,MTL2,MTL3}. Most approaches often incorporate deliberate designs for task-sharing and task-specific components to enable effective joint prediction across tasks. If the three stages are forcibly merged into a single training stage, two major issues will arise: 1) 
The model is trained with inputs that involve Random Modality Dropout, Random Modality Shuffle, and Random Spatial Masking. The encoder is supervised using the NT-Xent contrastive loss, an adapter is incorporated, and the model is simultaneously tasked with both segmentation and reconstruction. Such a fully entangled training setup lacks a clear task hierarchy, and there is no explicit coordination among the input, encoder, decoder, and output. 
2) 
When all tasks are treated equally without ordering, joint optimization becomes highly difficult. In the unified single-stage training attempt, the total loss involves six parts.  
As shown in Tab.~\ref{tab:ablation_study}, we compare the three-stage model with the unified single-stage model.
We observed two phenomena during the experiments: 1) 
The single stage model failed to converge, with loss plateauing early.
 2) The optimization process was highly unstable, with different losses fluctuating in turn and failing to decrease consistently together. 
 These phenomena and performace clearly indicate that unified single stage training is unable to achieve effective coordination among the various designs and supervision signals. The lack of a clear training order, combined with competing optimization objectives, leads to mutual interference.

\noindent\textbf{Reverse Attention Adapter}. 
Tab.~\ref{tab:ablation_study_RAA} shows ablations on the reverse attention adapter.  There are three key findings: 1) Removing reverse attention alone leads to a 4.2\% average drop, confirming its role in semantic compensation for missing modalities. 2) Subsequent elimination of mutual attention yields negligible performance variation, indicating that only emphasizing high-response regions between adapter layers and frozen encoding layers cannot achieve meaningful feature compensation. 3) Replacing the 3D Swin Transformer with a 3D convolution of similar parameters significantly degrades performance, validating the Transformer’s advantage in cross-modal correlation modeling. Additionally, we compare freezing vs. fine-tuning the encoder during stage 3. Fine-tuning causes over 6.4\% performance drops, indicating that it destroys the adapter-encoder synergy and the compensation mechanism relies on stable encoder representations rather than task-specific tuning.

To further explain this observation, we provide two perspectives: the inheritance between Stage~2 and Stage~3, and the rationale behind the lightweight reverse attention adapter (RAA).  
1) Unlike general SSL methods~\cite{SimCLR,MoCo,MAE}  that focus on learning generic representations without targeting specific downstream tasks, our contrastive learning in Stage~2 is task-aware.  We co-train the segmentation head to guide the encoder toward modality-invariant features that are directly beneficial for segmentation, rather than general-purpose representations. This task-guided design ensures that the learned contrastive space aligns with the downstream segmentation objective. Therefore, fine-tuning the encoder in Stage~3 would undermine the task-guided contrastive representations that were carefully established.  
2) The RAA itself is designed as a residual correction bridging the encoder representation gap between incomplete and complete modality inputs. Formally:  
\begin{equation}
f_{\text{inc}} + \mathcal{A}(f_{\text{inc}}) \approx f_{\text{com}},
\end{equation}
where $f_{\text{inc}}$ denotes encoder features from incomplete modalities, $f_{\text{com}}$ represents encoder features from complete modalities, and $\mathcal{A}(\cdot)$ is the learnable adapter. During Stage~3, both $f_{\text{inc}}$ and $f_{\text{com}}$ are frozen, while only $\mathcal{A}$ is trained. This constrained setup ensures that the adapter focuses purely on compensating missing information, enabling stable and efficient optimization. Once the encoder is unfrozen, however, all three components become variables, making their roles unclear and the optimization unstable. In conclusion, freezing the encoder in Stage~3 is a deliberate and necessary design choice to preserve the task-guided contrastive representations learned earlier. The RAA thus works in synergy with a stable encoder to effectively compensate for missing modalities and maintain robust segmentation performance.

\noindent\textbf{Hierarchical Compensation Mechanism}.  
As shown in Tab.~\ref{tab:ablation_study_three_level}, single-level compensations each surpass the baseline. Cross-level combinations reveal nonlinear synergy. Notably, merely integrating input- and feature-level compensations outperforms  existing methods in Tab.~\ref{tab:sota_compare}. 
The fully compensated model achieves 36.2\% average gain over baseline across categories, demonstrating consistent collaborative synergy among all three levels without mutual exclusion. To quantify modality representation gaps, Fig.~\ref{fig:gap_value} visualizes the average feature activation values across different encoding layers under various modality combinations. Fig.~\ref{fig:gap_value_1} displays the baseline visual distributions, where single-modality, dual-modality, and triple-modality combinations exhibit pronounced representation gaps compared to the full-modality setup. Fig.~\ref{fig:gap_value_2} shows results after stage 1 pre-training, where the compensation capability for single-modality inputs is significantly enhanced. Fig.~\ref{fig:gap_value_3} illustrates the results under the joint constraints of contrastive learning and segmentation tasks, where the representations of all the modalities are further aligned. In Fig.~\ref{fig:gap_value_4}, the reverse attention adapter in stage 3 effectively compensates for cumulative errors from modality reconstruction in stage 1 and spatial discrepancy differentiation in stage 2, using only a small number of parameters.

\section{Conclusion}
In this work, we propose a novel modality-relax segmentation framework based on a hierarchical self-supervised compensation strategy. 
Through multi-granularity data perturbation, segmentation task-guided feature distance constraints, and the design of a reverse attention adapter, we simultaneously integrate three different self-supervised techniques into the proposed UniMRSeg model and complete the representation compensation of the missing modality from the input level, feature level, and output level.
UniMRSeg is simple yet effective, achieving dominant performance in four different multi-modal image segmentation tasks. We hope that this research paradigm focusing on representation-level compensation can inspire more visual tasks that require modality-relax conditions in the future.

\bibliographystyle{abbrv}
\bibliography{main}
\newpage

\appendix
\section*{Appendix}
\label{sec:Appendix}

\begin{table*}[!t]
	\centering
    \caption{
    Quantitative comparison of brain tumor segmentation on BraTS2020. $\uparrow$ and $\downarrow$ indicate that higher and lower scores are better, respectively. 
\textbf{Note:} The reported results of RFNet~\cite{RFNet} and mmFormer~\cite{mmformer} are taken from the PASSION paper~\cite{PASSION}, where they are evaluated using the same training and test splits of BraTS2020.
	}
    \setlength{\abovecaptionskip}{2pt}
    \renewcommand{\arraystretch}{1.1}
    \renewcommand{\tabcolsep}{0.5mm}
	\resizebox{\linewidth}{!}{\begin{tabular}{cccc||cccc|cccc|cccc}
	\hline
	\multicolumn{4}{c||}{} & \multicolumn{12}{c}{{Dice score(\%)}} \\ \cline{5-16} 
	\multicolumn{4}{c||}{\multirow{-2}{*}{{Modality}}} & \multicolumn{4}{c|}{{Whole Tumor (Whole)}} & \multicolumn{4}{c|}{{Tumor Core (Core)}} & \multicolumn{4}{c}{{Enhancing Tumor (Enhancing)}} \\
	\hline
	
	\multirow{2}{*}{Flair} & \multirow{2}{*}{{T1}} & \multirow{2}{*}{{T1ce}} & \multirow{2}{*}{{T2}} & {{RFNet}} & {{mmFormer}} & {{PASSION-RFNet}} & {{UniMRSeg}} & 
    {{RFNet}} & {{mmFormer}} & {{PASSION-RFNet}} & {{UniMRSeg}} & 
  {{RFNet}} & {{mmFormer}} & {{PASSION-RFNet}} & {{UniMRSeg}}   \\
   &&&& {{~\cite{RFNet}}} & {{~\cite{mmformer}}} & {{~\cite{PASSION}}}  & {{}} &  {{~\cite{RFNet}}} & {{~\cite{mmformer}}} & {{~\cite{PASSION}}}  & {{}} &  {{~\cite{RFNet}}} & {{~\cite{mmformer}}} & {{~\cite{PASSION}}}  & {{}} \\
	\hline
	\textbf{$\circ$} & \textbf{$\circ$} & \textbf{$\circ$} & \textbf{$\bullet$} 
    & 82.25& 83.33&  81.35&  \textbf{84.54} & 66.77& 63.74& 53.69& \textbf{68.94} &40.08& 38.20& 32.13& \textbf{47.41} \\
	\textbf{$\circ$} & \textbf{$\circ$} & \textbf{$\bullet$} & \textbf{$\circ$} & 69.27&  69.87&  75.02&  \textbf{80.49} & 77.24& 76.40& 77.25 & \textbf{78.33} &  65.81&  67.25& 69.15 & \textbf{68.79} \\
	\textbf{$\circ$} & \textbf{$\bullet$} & \textbf{$\circ$} & \textbf{$\circ$} &68.53&  69.60&  72.20 & \textbf{79.72} & 59.53&56.00& 57.20& \textbf{63.32} & 31.99& 28.66& 30.42 & \textbf{40.12} \\
	\textbf{$\bullet$} & \textbf{$\circ$} & \textbf{$\circ$} & \textbf{$\circ$} &  82.28&  83.34&  83.95 & \textbf{84.45} & 64.30& 61.74& 58.60& \textbf{66.69} & 36.67& 33.76& 26.40& \textbf{41.17} \\
	\textbf{$\circ$} & \textbf{$\circ$} & \textbf{$\bullet$} & \textbf{$\bullet$} & 83.94&  85.47&  82.60 & \textbf{85.66} & 81.90& 81.36& 78.16& \textbf{82.43} & 69.18& 69.83& 69.52 & \textbf{71.48} \\
	\textbf{$\circ$} & \textbf{$\bullet$} & \textbf{$\bullet$} & \textbf{$\circ$} & 74.07&  74.74&  76.45 & \textbf{81.46} & 81.45& 80.50& 79.83 & \textbf{82.06} & 68.56& 70.70& 70.34 & \textbf{71.38} \\
	\textbf{$\bullet$} & \textbf{$\bullet$} & \textbf{$\circ$} & \textbf{$\circ$} & 85.51&  86.60& \textbf{86.78}& {86.27} & 70.28& 67.07& 65.57& \textbf{72.00} & 41.19& 38.51& 36.62 & \textbf{48.45} \\
	\textbf{$\circ$} & \textbf{$\bullet$} & \textbf{$\circ$} & \textbf{$\bullet$} & 84.90&  85.81&  82.30& \textbf{85.97} & 70.39& 66.61& 61.70& \textbf{72.09} & 43.61& 41.02& 37.26&  \textbf{54.44} \\
	\textbf{$\bullet$} & \textbf{$\circ$} & \textbf{$\circ$} & \textbf{$\bullet$} & 86.42&  \textbf{87.73}&  86.85& {86.76} & 70.70& 68.66& 61.54& \textbf{72.34} & 43.62& 42.17& 31.47&  \textbf{56.44} \\
	\textbf{$\bullet$} & \textbf{$\circ$} & \textbf{$\bullet$} & \textbf{$\circ$} & 85.18&  \textbf{87.39}&  85.57&  {86.54} & 80.16& 79.67& 78.38& \textbf{81.91} & 68.38& 68.11& 68.60& \textbf{71.83} \\
	\textbf{$\bullet$} & \textbf{$\bullet$} & \textbf{$\bullet$} & \textbf{$\circ$} & 86.61&  87.99&  87.46&  \textbf{88.81} & 81.67& 80.71& 79.54& \textbf{82.41} & 69.47& 70.86& 72.24&  \textbf{74.93} \\
	\textbf{$\bullet$} & \textbf{$\bullet$} & \textbf{$\circ$} & \textbf{$\bullet$} & 87.63&  88.60&  87.91 & \textbf{89.87} & 72.83& 69.89& 66.50& \textbf{80.69} & 45.21& 43.22& 37.48& \textbf{53.49} \\
	\textbf{$\bullet$} & \textbf{$\circ$} & \textbf{$\bullet$} & \textbf{$\bullet$} & 87.36&  88.84&  87.45 & \textbf{89.46} & 81.97& 81.25& 78.88& \textbf{83.68} & 68.76& 69.91& 70.18&  \textbf{74.37} \\
	\textbf{$\circ$} & \textbf{$\bullet$} & \textbf{$\bullet$} & \textbf{$\bullet$} & 85.40&  86.40&  83.04&   \textbf{89.04} & 83.28& 82.23& 80.00& \textbf{84.64} & 71.04& 70.82& 69.90& \textbf{74.56} \\
	\textbf{$\bullet$} & \textbf{$\bullet$} & \textbf{$\bullet$} & \textbf{$\bullet$} & 88.30&  89.27&  88.20&   \textbf{90.45} & 82.80& 81.90& 80.79 & \textbf{85.26} & 69.64& 70.62& 71.35&  \textbf{75.96} \\
	\hline
	\multicolumn{4}{c||}{{Average $\uparrow$}} & 82.51 & 83.67 & 83.15 & \textbf{85.97} & 75.02  & 73.18 & 70.51 & \textbf{77.12} & 55.55 & 54.91 & 52.87 & \textbf{61.65} \\
	\hline
	\multicolumn{4}{c||}{{Std Dev $\downarrow$}}  & 6.27 & 6.45 & 4.86 & \textbf{3.26}  & 7.45 & 8.48 & 9.66 & \textbf{6.95}  & 14.56 & 16.24 & 18.69 & \textbf{12.81} \\
	\hline
\end{tabular}}
	
	\label{tab:sota_brats2020_rfnet}
\end{table*}

\begin{table*}[!t]
	\centering
    \caption{
  Quantitative comparison of brain tumor segmentation on BraTS2018. $\uparrow$ and $\downarrow$ indicate that higher and lower scores are better, respectively. 
\textbf{Note:} The reported results of mmFormer~\cite{mmformer}, M3AE~\cite{M3AE} and M3FeCon~\cite{zeng2024missing} are taken from the M3FeCon~\cite{zeng2024missing}, where they are evaluated using the same training and test splits of BraTS2018.
	}
    \setlength{\abovecaptionskip}{2pt}
    \renewcommand{\arraystretch}{1.1}
    \renewcommand{\tabcolsep}{0.5mm}
	\resizebox{\linewidth}{!}{\begin{tabular}{cccc||cccc|cccc|cccc}
	\hline
	\multicolumn{4}{c||}{} & \multicolumn{12}{c}{{Dice score(\%)}} \\ \cline{5-16} 
	\multicolumn{4}{c||}{\multirow{-2}{*}{{Modality}}} & \multicolumn{4}{c|}{{Whole Tumor (Whole)}} & \multicolumn{4}{c|}{{Tumor Core (Core)}} & \multicolumn{4}{c}{{Enhancing Tumor (Enhancing)}} \\
	\hline
	
	\multirow{2}{*}{Flair} & \multirow{2}{*}{{T1}} & \multirow{2}{*}{{T1ce}} & \multirow{2}{*}{{T2}} & {{mmFormer}} & {{M3AE}} & {{M3FeCon}} & {{UniMRSeg}} & {{mmFormer}} & {{M3AE}} & {{M3FeCon}} & {{UniMRSeg}} &  {{mmFormer}} & {{M3AE}} & {{M3FeCon}} & {{UniMRSeg}}  \\
   &&&& {{~\cite{mmformer}}} & {{~\cite{M3AE}}} & {{~\cite{zeng2024missing}}} & {{}} &  {{~\cite{mmformer}}} & {{~\cite{M3AE}}} & {{~\cite{zeng2024missing}}} & {{}} &  {{~\cite{mmformer}}} & {{~\cite{M3AE}}} & {{~\cite{zeng2024missing}}} & {{}} \\
	\hline
	\textbf{$\circ$} & \textbf{$\circ$} & \textbf{$\circ$} & \textbf{$\bullet$} & 81.43& 84.22& 85.13& \textbf{87.10}& 64.61& 69.14& 72.48& \textbf{76.45}& 41.92& 46.93& 49.32&\textbf{54.20} \\
	\textbf{$\circ$} & \textbf{$\circ$} & \textbf{$\bullet$} & \textbf{$\circ$} & 72.62& 75.16& 75.26& \textbf{82.22}& 75.93& \textbf{82.53}& 82.31& {82.25}& 71.37& 73.04& \textbf{75.88}&{75.46} \\
	\textbf{$\circ$} & \textbf{$\bullet$} & \textbf{$\circ$} & \textbf{$\circ$} & 67.92& 73.83& 75.24& \textbf{82.29}& 56.96& 65.77& 65.72& \textbf{69.70}& 31.38& 36.54& 44.35&\textbf{49.20} \\
	\textbf{$\bullet$} & \textbf{$\circ$} & \textbf{$\circ$} & \textbf{$\circ$} & 86.37& 88.04& 89.05& \textbf{89.85}& 61.61& 66.02& 69.42& \textbf{73.52}& 37.98& 34.96& 46.59&\textbf{51.35} \\
	\textbf{$\circ$} & \textbf{$\circ$} & \textbf{$\bullet$} & \textbf{$\bullet$} & 83.25& 85.58& 86.61& \textbf{88.57}& 79.07& 83.85& 84.75& \textbf{84.85}& 73.13& 74.39& 76.55&\textbf{76.75} \\
	\textbf{$\circ$} & \textbf{$\bullet$} & \textbf{$\bullet$} & \textbf{$\circ$} & 74.75& 76.50& 79.16& \textbf{83.25}& 79.01& 83.11& 82.88& \textbf{83.97}& 72.77& 74.58& 76.78&\textbf{78.41} \\
	\textbf{$\bullet$} & \textbf{$\bullet$} & \textbf{$\circ$} & \textbf{$\circ$} & 87.46& 88.49& 90.17& \textbf{91.45}& 66.47& 70.53& 72.82& \textbf{76.72}& 41.75& 48.16& 48.47&\textbf{53.20} \\
	\textbf{$\circ$} & \textbf{$\bullet$} & \textbf{$\circ$} & \textbf{$\bullet$} & 82.46& 86.34& 86.03& \textbf{88.15}& 69.89& 71.46& 72.01& \textbf{77.43}& 43.85& 44.73& 50.17&\textbf{55.12} \\
	\textbf{$\bullet$} & \textbf{$\circ$} & \textbf{$\circ$} & \textbf{$\bullet$} & 87.99& 89.31& 90.46& \textbf{91.78}& 70.18& 70.59& 72.81& \textbf{76.95}& 46.47& 40.57& 51.12&\textbf{56.09} \\
	\textbf{$\bullet$} & \textbf{$\circ$} & \textbf{$\bullet$} & \textbf{$\circ$} & 87.61& 88.85& 90.35& \textbf{91.98}& 78.34& 84.03& 84.14& \textbf{85.20}& 73.93& 74.08& 76.23&\textbf{78.43} \\
	\textbf{$\bullet$} & \textbf{$\bullet$} & \textbf{$\bullet$} & \textbf{$\circ$} & 87.77& 88.07& 89.62& \textbf{89.99}& 80.22& 83.72& 84.81& \textbf{85.41}& 74.31& 73.42& 76.74& \textbf{78.72} \\
	\textbf{$\bullet$} & \textbf{$\bullet$} & \textbf{$\circ$} & \textbf{$\bullet$} &88.08& 89.24& 90.24& \textbf{91.89}& 71.97& 72.41& 75.98& \textbf{80.26}& 46.51& 44.15& 52.63& \textbf{60.12} \\
	\textbf{$\bullet$} & \textbf{$\circ$} & \textbf{$\bullet$} & \textbf{$\bullet$} &88.47& 89.46& 90.07& \textbf{92.01}& 79.94& 84.25& 84.64& \textbf{85.47}& 74.53& 74.68& 77.64& \textbf{79.02} \\
	\textbf{$\circ$} & \textbf{$\bullet$} & \textbf{$\bullet$} & \textbf{$\bullet$} & 83.08& 85.06& 86.79& \textbf{90.48}& 80.89& 84.13& 84.39& \textbf{85.49}& 73.49 & 73.37& 77.33& \textbf{79.36} \\
	\textbf{$\bullet$} & \textbf{$\bullet$} & \textbf{$\bullet$} & \textbf{$\bullet$} & 89.93& 89.56& 90.69& \textbf{92.45}& 86.23& 84.24& 85.03& \textbf{86.44}& 76.36& 74.85& 77.81& \textbf{79.87} \\
	\hline
	\multicolumn{4}{c||}{{Average $\uparrow$}} &83.28 & 85.18& 86.32 & \textbf{88.90} & 73.42 & {77.05} & 78.28 & \textbf{80.67} & 58.65 & 59.23 & {63.84} & \textbf{67.02} \\
	\hline
	\multicolumn{4}{c||}{{Std Dev $\downarrow$}} & 6.37 &  5.29 &  5.25& \textbf{3.50} & 8.02 & 7.34 & 6.60 & \textbf{5.05} & 16.51 & 16.17 & 14.05 & \textbf{12.25} \\
	\hline
\end{tabular}}
	
	\label{tab:sota_compare_brats2018}
\end{table*}

\section{Performance Comparison on BraTS2020 and BraTS2018}
To enable fair comparisons with a broader range of methods, we adopt the same data split settings as used in the PASSION paper~\cite{PASSION} for BraTS2020, where the dataset is divided into 219 cases for training, 50 for validation, and 100 for testing. For BraTS2018, we follow the data split protocol from M3FeCon~\cite{zeng2024missing}, using 200 cases for training and 85 for testing. As shown in Tab.~\ref{tab:sota_brats2020_rfnet} and Tab.~\ref{tab:sota_compare_brats2018}, our method consistently achieves the best average performance and lowest standard deviation.

\section{Necessity of the Three-Stage Training Strategy}
The proposed three-stage training pipeline, comprising Multi-granular Modality Reconstruction, Modality-Invariant Contrastive Learning, and Incomplete Modality Adaptive Fine-tuning, is designed to address the unique challenges of missing modality segmentation, which cannot be effectively tackled using standard end-to-end training.

Unlike single-modality RGB image tasks that benefit from large-scale pre-trained models, many other modalities such as depth, infrared, and medical imaging still lack general-purpose pre-trained encoders. As a result, most models initialized randomly suffer from suboptimal feature representations and unstable performance. Moreover, simple end-to-end optimization struggles to bridge the representation gap between complete and incomplete modality inputs, leading to poor generalization to unseen modality combinations.

To this end, each stage in our training strategy plays a complementary role in strengthening the model's robustness and adaptability:
\begin{itemize}[leftmargin=*,itemsep=0em,topsep=0em,parsep=0em]
\item Stage 1 (Modality Reconstruction): Enables the model to learn cross-modality structural priors and recover semantically meaningful representations from incomplete inputs.
\item Stage 2 (Contrastive Learning): Encourages the network to align features between complete and missing modality inputs, enhancing modality-invariant representations.
\item Stage 3 (Unified Fine-tuning): Refines prediction consistency across different modality combinations and ensures deployment under a unified model.
\item Each stage serves a distinct and complementary role (input-level structure recovery, feature-level modality alignment, and output-level consistency). Extensive ablation studies (Sec.~\ref{sec:Ablation}) validate that each stage contributes distinct improvements in average accuracy. Without any of the stages, performance consistently degrade. 
\end{itemize}

\section{Discussion on Training Complexity and Efficiency}
\label{sec:Efficiency}
Although UniMRSeg adopts a three-stage training strategy, it is deliberately designed for efficiency and simplicity. We analyze the training complexity from two perspectives: concise design and inference simplicity.

\textbf{Concise Design:}
All three stages in UniMRSeg share a unified 3D U-Net-style encoder-decoder backbone, integrated with a lightweight 3D ASPP module. No additional complex or stacked modules are introduced. This ensures both architectural clarity and training efficiency. Stage 1 utilizes a masked autoencoding-based reconstruction head. This head operates on top of the shared encoder and does not require extra encoder-decoder branches, keeping the design compact. Stage 2 applies a contrastive loss to already-computed latent features, without introducing new network components. Stage 3 focuses on refining segmentation via a lightweight reverse attention adapter and decoder. The encoder is frozen, making this phase computationally efficient and requiring minimal fine-tuning.

\textbf{Single-Stage Inference with Unified Weights:} 

\begin{wraptable}{r}{0.45\textwidth}
  \centering
    \caption{Efficiency comparison.}
  \renewcommand{\arraystretch}{1.1}
    \renewcommand{\tabcolsep}{0.5mm}
	\resizebox{\linewidth}{!}{	\begin{tabular}{l||cccc}
	\hline                                                            Metrics & RFNet~\cite{RFNet}&mmFormer~\cite{mmformer}&M3AE~\cite{M3AE}&UniMRSeg \\                                                                                                                           
	\hline
		{{Parameters (MB)$\downarrow$}}  & {{{34}}} & {{106}} & {{167}} & {{87}} \\
		{{FLOPs (G)$\downarrow$}}  & {{148}} & {{748}} & {{248}} & {{202}} \\

	\hline
\end{tabular}

}
    \label{tab:efficiency_comparison}
\end{wraptable}
Despite the multi-stage training, inference remains a single-stage process. 
The final model trained at stage 3 merges all learned representations into a unified network with shared parameters across all modality combinations. Unlike other methods that require ensemble inference~\cite{M3AE,KD_brats} or modality-specific branches~\cite{zeng2024missing,mmformer}, UniMRSeg performs fast, unified inference without requiring modality recognition or dynamic model selection. As shown in Tab.~\ref{tab:efficiency_comparison}, UniMRSeg achieves comparable computational efficiency to state-of-the-art methods.

\section{Modality-Agnostic Robustness via Random Modality Shuffle}
Previous multi-modal segmentation methods~\cite{zeng2024missing,mmformer,RFNet,M3AE} often rely on a fixed modality-to-channel correspondence, where each input modality (\eg, T1, T2, T1ce, Flari) is assigned to a specific encoder or channel during both training and inference. This design imposes a strong prior assumption: the modality type of each input channel must be known in advance and must strictly align with the model's expected input structure during inference. However, such assumptions significantly limit the scalability and automation of AI-driven medical image analysis pipelines, especially in real-world applications where modality labels may be missing, inconsistent, or ambiguous.

To alleviate this constraint, we introduce a random modality shuffle strategy in stage 1, where the input modalities are randomly permuted at each training iteration. This forces the model to learn modality-invariant representations and reduces its reliance on fixed input orders.

\begin{wraptable}{r}{0.6\textwidth}
  \centering
    \caption{Ablation study on the impact of different modality-channel orders during inference. 
\texttt{Flair-T1-T1ce-T2} denotes the best-performing fixed input order, while \texttt{5-Ave} represents the average result of five inference runs using random modality shuffle. }
  \renewcommand{\arraystretch}{1.1}
    \renewcommand{\tabcolsep}{0.5mm}
	\resizebox{0.7\linewidth}{!}{	\begin{tabular}{l||ccc}
	\hline                                                                                   & \multicolumn{3}{c}{{Average Dice score(\%)}}                                                                                                                                                                                                                                                                                                                                                                                                                                                                                                                                                                                         \\ \cline{2-4} 
	\multicolumn{1}{l||}{\multirow{-2}{*}{{Models}}}                                                                                               & {{Whole}}                                                                                                                                                                        & {{Core}}                                                                                                                                                                         & {{Enhancing}}                                                                                                                                                                     \\
	\hline
		\texttt{{Flair-T1-T1ce-T2}}  & {{80.64}} & {{73.33}} & {{63.10}} \\
		\texttt{{5-Ave}}  & {{80.54}} & {{73.27}} & {{63.05}} \\

	\hline
\end{tabular}

}
\label{tab:ablation_random_shuffle}
\end{wraptable}
As shown in Tab.~\ref{tab:ablation_random_shuffle}, even after randomly shuffling the modality-channel order five times during inference, the segmentation performance remains stable with minimal variation. These results demonstrate that the random shuffle not only improves performance (see Tab.~\ref{tab:ablation_study}) but also enhances practical robustness and scalability. This aligns with the broader goal of building fully automated and generalizable medical AI systems.

\section{Limitations and Future Work}
\label{app:sec:Limitations}
While the three-stage training strategy significantly improves model generalizability and segmentation accuracy under missing modalities, it inevitably increases the training pipeline complexity. This additional training overhead may pose challenges for practitioners in time-constrained or resource-limited environments. We acknowledge this trade-off and consider streamlining or accelerating the training process as a critical future direction. Several promising avenues to achieve this include: \noindent\textit{\textbf{\uppercase\expandafter{\romannumeral1})
Curriculum-based Optimization.}}  Introducing a curriculum learning schedule that gradually transitions from reconstruction to segmentation may allow for progressive learning in a single stage. 
\noindent\textit{\textbf{\uppercase\expandafter{\romannumeral2})
Modality-aware Parameter Sharing.}} 
 Parameter-efficient fine-tuning (\eg, adapters or LoRA) across stages may allow most model weights to be reused, reducing memory and time overhead. 
\noindent\textit{\textbf{\uppercase\expandafter{\romannumeral3})
Cross-task Shared Training Objectives.}}  
 Reformulating the three tasks under a shared objective (\eg, information bottleneck or consistency maximization) may unify their gradients and reduce stage-specific training routines. These directions hold the potential to preserve the benefits of multi-stage training while simplifying the optimization process, making the proposed method more accessible to broader applications in multi-modal image analysis.

\end{document}